\documentclass[pdflatex,sn-mathphys-num]{sn-jnl}

\usepackage{graphicx}%
\usepackage{multirow}%
\usepackage{amsmath,amssymb,amsfonts}%
\usepackage{amsthm}%
\usepackage{mathrsfs}%
\usepackage[title]{appendix}%
\usepackage{xcolor}%
\usepackage{textcomp}%
\usepackage{manyfoot}%
\usepackage{booktabs}%
\usepackage{tcolorbox}
\usepackage{algorithm}%
\usepackage{algorithmicx}%
\usepackage{algpseudocode}%
\usepackage{listings}%
\usepackage{tabularx}
\usepackage{booktabs}
\usepackage{array}
\usepackage{colortbl}
\usepackage{makecell}
\usepackage{pifont}
\usepackage{amssymb}

\definecolor{headerblue}{RGB}{235,242,250}
\definecolor{dasblue}{RGB}{220,234,247}
\definecolor{rowgray}{RGB}{248,248,248}

\newcommand{\cmark}{\ding{51}}   
\newcommand{\xmark}{\ding{55}}   

\tcbuselibrary{skins, breakable}

\newtcolorbox[auto counter, number within=section]{clinicalbox}[2][]{
  colback=gray!5!white,
  colframe=gray!75!black,
  fonttitle=\bfseries,
  title={Box~\thetcbcounter: #2},
  label={#1}
}

\theoremstyle{thmstyleone}%
\theoremstyle{thmstyletwo}%

\theoremstyle{thmstylethree}%

\begin{document}

\title[Article Title]{Beyond Federated Learning: Decentralized Agentic Systems for Self-Evolving Clinical Intelligence}



\author*{\fnm{Haoyuan} \sur{Li}}\email{h.y.li@tue.nl}

\author{\fnm{Mathias} \sur{Funk}}\email{M.Funk@tue.nl}

\author{\fnm{Aaqib} \sur{Saeed}}\email{a.saeed@tue.nl}

\affil{\orgname{Eindhoven University of Technology}, \orgaddress{\city{Eindhoven}, \country{The Netherlands}}}


\abstract{Clinical expertise is inherently distributed: it resides not in any single dataset or institution but in the accumulated reasoning experiences of thousands of clinicians navigating diverse patient populations, resource constraints, and evolving treatment protocols. Yet current approaches to scaling medical AI, from centralized large language models to federated learning, remain anchored to a data-centric paradigm that treats knowledge as static parameters to be aggregated. In this Perspective, we argue for a fundamental shift from \emph{data aggregation} to \emph{experience aggregation}. We propose Decentralized Agentic Systems (DAS), a framework in which distributed AI agents operating across healthcare institutions synthesize collective clinical intelligence by sharing abstracted reasoning trajectories rather than raw data or model parameters. Unlike federated learning, which averages numerical gradients across homogeneous model architectures, DAS exchanges structured records of clinical reasoning, specifically the sequence of observations, hypotheses, tool invocations, and outcome feedback that constitute diagnostic experience. We examine the architectural principles, deployment considerations, governance requirements, and critical open challenges of this framework, laying the groundwork for a collaborative medical AI ecosystem where distributed expertise autonomously adapts to the complexities of longitudinal patient care.}




\maketitle

\section{Introduction}

Medicine is, at its core, a distributed enterprise. The expertise required to manage a complex patient rarely resides in a single clinician, department, or institution. A rural primary care physician develops pattern recognition for atypical presentations under resource constraints that an academic specialist never encounters; a community cardiologist accumulates experience with medication titration across diverse socioeconomic contexts; a home health monitoring system learns to distinguish clinically significant decompensation from benign fluctuations in daily vital signs. This distributed expertise, forged through thousands of diagnostic encounters across heterogeneous clinical environments, represents the most valuable and least utilized knowledge asset in healthcare.

The rapid development of Large Language Models (LLMs) has contributed to growing interest in agentic Artificial Intelligence (AI), which seeks to extend AI systems beyond static information retrieval toward perception, reasoning, and action in dynamic environments. Prior work has reported promising results in complex domains such as software engineering, mathematical theorem proving, and scientific hypothesis generation \cite{swe-agent, prover, ai-coscientist, cuda-engineer}. In healthcare, where clinical reasoning demands synthesizing heterogeneous evidence under uncertainty, agentic AI promises to augment human expertise by orchestrating knowledge-intensive diagnostic and therapeutic workflows.

Early applications of agentic AI in medicine established foundational approaches to algorithmic clinical reasoning. Systems such as those in \cite{oncology, Medagents, meddxagent, personalizedagent} employ LLMs as centralized orchestrators, using modular decomposition to partition clinical queries into sub-tasks spanning evidence retrieval, differential diagnosis, and treatment recommendation. While effective for localized decision support, these systems operate as isolated instances constrained by static reasoning templates and single-institution deployment. Recent work has advanced toward hierarchical multi-agent systems enabling dynamic collaboration for symptom assessment, diagnostic interpretation, and treatment recommendation, simulating multi-turn clinical consultations \cite{medagentsim, cardiovascularagent, echocardiography, mdagents, care-ad}.

However, a fundamental architectural tension pervades current systems: clinical expertise is inherently distributed across institutions, specialties, and care settings, yet agentic AI remains stubbornly centralized. This tension is deeper than the well-known challenges of data fragmentation and privacy regulation, though it encompasses them. We identify three dimensions of this mismatch, each reflecting not merely a data problem but an \emph{expertise} problem.

\paragraph{Expertise fragmentation problem.} The conventional framing of healthcare data silos emphasizes regulatory barriers to data sharing. Though real, this framing obscures a more fundamental issue: even removing regulatory barriers would not pool clinical \emph{expertise}. The diagnostic insight a rural emergency physician develops managing acute coronary syndrome without catheterization access is not encoded in the EHR; it resides in the physician's reasoning process and outcomes observed. A centralized system trained on pooled records acquires statistical patterns but not the situated reasoning constituting clinical mastery. The challenge is therefore not merely to share data across institutions but to share the \emph{experience} of navigating clinical complexity.

\paragraph{Scaling limits of static knowledge.} The advancement of LLMs has been predicated on access to ever-larger corpora of human-generated text, yet recent analyses suggest this paradigm is approaching asymptotic limits as high-quality medical data becomes increasingly scarce \cite{positiondata, scalingdata}. In healthcare, where rare conditions, evolving treatment protocols, and longitudinal patient trajectories demand continuous learning, reliance on static datasets creates systems that are perpetually outdated. More critically, static training data captures \emph{declarative} medical knowledge such as facts, guidelines, and associations, but not the \emph{procedural} knowledge of how to apply it under the uncertainty and time pressure of clinical practice.

\paragraph{Sovereignty-utility tension.} Autonomous agents handling Protected Health Information (PHI) must comply with HIPAA, IVDR, and GDPR \cite{hipaa, ivdr, gdpr}, yet compliance is only one dimension of a broader sovereignty challenge. Healthcare institutions increasingly assert governance over not only their data but their institutional knowledge (i.e., clinical protocols, reasoning conventions, and accumulated expertise) as competitive and intellectual assets. Current architectures offer institutions a binary choice: contribute data to centralized systems or remain isolated. What is needed is an architecture that enables institutions to \emph{contribute to and benefit from collective intelligence while retaining sovereignty over their clinical assets}, sharing insights without surrendering the experiences from which those insights derive.

These challenges are compounded by the computational realities of deployment. State-of-the-art agents rely on multi-model toolchains: EchoAgent \cite{echoagent} orchestrates domain-specific vision models for echocardiogram interpretation, while Echo-CoPilot \cite{echocopilot} leverages EchoPrime \cite{echoprime} and PanEcho \cite{panecho} for multi-view cardiac analysis. Centralizing such systems incurs substantial inference costs \cite{agentcontexteng}, while naive distribution faces severe infrastructure heterogeneity.

Such converging pressures cannot be addressed by existing distributed learning paradigms. Federated Learning (FL), the prevailing privacy-preserving approach, aggregates model parameters through gradient averaging, operating at the wrong level of abstraction for clinical expertise. Gradients encode statistical correlations rather than reasoning strategies, require architectural homogeneity, and provide no mechanism for transferring the procedural knowledge characterizing clinical mastery.

In this Perspective, we propose Decentralized Agentic Systems (DAS) as an alternative paradigm: a framework wherein distributed agents operating across healthcare institutions synthesize collective clinical intelligence through the sharing of abstracted reasoning trajectories, while concurrently self-evolving through the reciprocal incorporation of network-wide experiential knowledge into local reasoning capabilities. We formalize the architectural principles underlying DAS (Section~\ref{sec:arch_das}), examine the deployment landscape and governance requirements for real-world adoption (Sections~\ref{sec:deploy} and~\ref{sec:govern}), and articulate the critical open challenges (Section~\ref{sec:challenges}).
\section{Learning from Experience Beyond Human Data}

The transition toward decentralized agentic architectures represents a necessary evolution aligned with fundamental shifts in how AI acquires knowledge. LLMs' progress has been predicated on scaling human-curated datasets, a paradigm approaching asymptotic limits as high-quality clinical data grows scarce \cite{positiondata, scalingdata}. In healthcare, this constraint is particularly acute: medical knowledge is continuously generated through diagnostic encounters, treatment adaptations, and clinical judgments across thousands of institutions, accessible only in part to any centralized system. We argue that the next frontier lies in enabling AI agents to learn directly from distributed clinical experiences, transforming ephemeral diagnostic interactions into durable, shareable intelligence.

Experience differs fundamentally from data. Data represents static records, laboratory values, imaging findings, documented diagnoses. Experience, by contrast, encompasses the dynamic process through which an agent perceives clinical signals, reasons about their significance under uncertainty, selects among possible interventions, and observes outcomes \cite{experience}. Consider the difference between a dataset of electrocardiograms labeled with diagnoses and the experience of an agent that detects a subtle ST-elevation pattern, retrieves relevant patient history, considers confounding factors, and refines its interpretation based on downstream clinical confirmation. The former provides declarative knowledge, facts about the world; the latter generates procedural knowledge, that is, how to navigate clinical reasoning itself.

Each institutional agent continuously generates experiential feedback through clinical interactions. A cardiology-focused agent may develop nuanced expertise in distinguishing Takotsubo cardiomyopathy from acute coronary syndrome in post-menopausal women; a primary care agent may learn to recognize atypical thyroid dysfunction in elderly populations. Such experiences represent the contextual, procedural knowledge unavailable in static datasets. The challenge lies in synthesizing this distributed intelligence into collective clinical wisdom without violating data sovereignty constraints.

To address this, we propose DAS as a framework for agent-centric insight aggregation. Unlike FL \cite{fedlearning}, which aggregates numerical model parameters through gradient averaging, DAS operates at a higher abstraction level: the sharing of \textit{semantic trajectories}. We define a semantic trajectory as a structured record comprising an agent's reasoning path: observations, hypotheses, tool invocations, and inferential steps, together with environmental feedback that validated or refuted its conclusions. Exchanging these traces rather than raw data or model weights enables knowledge transfer that is both privacy-conscious and semantically meaningful, forming the substrate through which individual agentic nodes continuously self-evolve: each agent refines its local reasoning capabilities by assimilating experiential knowledge distilled from across the distributed network.

\paragraph{Why existing federated approaches are insufficient.} Gradient-based approaches assume homogeneity in model architectures, data distributions, and task definitions, while assumptions rarely hold across diverse healthcare settings. Semantic trajectories abstract away these implementation details to capture the core reasoning process. An insight such as ``in patients presenting with chest pain and elevated troponin but atypical ECG findings, consider requesting strain echocardiography before concluding myocardial infarction'' transfers meaningfully across institutional contexts regardless of underlying architectures or data distributions.

DAS is distinct from FL variants that partially address the limitations of FedAvg \cite{fedlearning}. Federated distillation \cite{lin2020ensemble, he2020group} exchanges soft predictions or logits, reducing architectural coupling. Split learning \cite{gupta2018distributed} partitions computation across clients and servers. Personalized FL \cite{t2020personalized, fallah2020personalized} accommodates heterogeneous data distributions through local adaptation. Each addresses a specific limitation of gradient-based aggregation, yet all share a common constraint: \emph{they operate at the level of model parameters or outputs, not at the level of reasoning}.

Table~\ref{tab:comparison} summarizes the key distinctions. The critical difference is representational: parameter-based approaches encode knowledge as numerical vectors requiring architectural compatibility. Semantic trajectories encode knowledge as structured reasoning sequences that are interpretable, auditable, and transferable across heterogeneous architectures. A trajectory recording reasoning from ambiguous ECG findings to a correct Wellens syndrome diagnosis transfers regardless of whether the receiving agent uses BioMistral, Med-PaLM, or a future architecture.

\begin{table}[t]
\centering
\caption{Comparison of distributed learning paradigms for healthcare AI. DAS operates at a 
fundamentally different level of abstraction than parameter-based approaches. 
Superscript~$^*$ denotes partial or limited applicability.}
\label{tab:comparison}
\setlength{\tabcolsep}{6pt}
\renewcommand{\arraystretch}{1.35}
\small
\begin{tabular}{@{} >{\raggedright\arraybackslash}p{3.0cm}
                    >{\centering\arraybackslash}p{2.0cm}
                    >{\centering\arraybackslash}p{2.2cm}
                    >{\centering\arraybackslash}p{2.1cm}
                    >{\centering\arraybackslash}p{2.1cm} @{}}
\toprule
\textbf{Property}
  & \textbf{FedAvg}
  & \textbf{Federated}\newline\textbf{Distillation}
  & \textbf{Personalized}\newline\textbf{FL}
  & \textbf{DAS (Ours)} \\
\midrule

Unit of sharing
  & Gradients / weights
  & Soft predictions / logits
  & Adapted parameters
  & \cellcolor{dasblue}Reasoning trajectories \\

\rowcolor{rowgray}
Model heterogeneity supported
  & \xmark
  & ~\xmark$^*$
  & \xmark
  & \cellcolor{dasblue}\cmark \\

Captures procedural reasoning
  & \xmark
  & \xmark
  & \xmark
  & \cellcolor{dasblue}\cmark \\

\rowcolor{rowgray}
Interpretability of shared knowledge
  & \xmark
  & ~\xmark$^*$
  & \xmark
  & \cellcolor{dasblue}~~~~~\cmark\newline(human-readable) \\

Cross-specialty transfer
  & \xmark
  & \xmark
  & ~\cmark$^*$
  & \cellcolor{dasblue}~~~~~\cmark\newline(semantic) \\

\rowcolor{rowgray}
Auditability for clinical safety
  & \xmark
  & \xmark
  & \xmark
  & \cellcolor{dasblue}~~~~~\cmark\newline(trajectory inspection) \\

Privacy risk vector
  & Gradient inversion
  & Output inversion
  & Gradient inversion
  & \cellcolor{dasblue}Semantic inference \\

\bottomrule
\end{tabular}
\end{table}

This comparison also highlights that DAS introduces its own privacy risk vector, namely semantic inference from reasoning patterns, which differs qualitatively from gradient inversion attacks and demands purpose-built mitigation strategies (Section~\ref{sec:privacy}).

\paragraph{Relationship to adjacent paradigms.} DAS relates to, but differs from, several active research directions. Retrieval-Augmented Generation (RAG) \cite{lewis2020retrieval} enhances LLM reasoning by retrieving relevant documents; DAS can be viewed as a federated extension of RAG where the retrieval corpus consists of validated reasoning trajectories from distributed agents. Chain-of-thought distillation \cite{ho2023large, magister2023teaching} transfers reasoning via training on reasoning traces; DAS extends this from a teacher-student relationship to a network of peer agents sharing experiential knowledge. Experience replay in multi-agent reinforcement learning \cite{foerster2017stabilising} enables learning from stored interactions; DAS adapts this to healthcare where experiences comprise clinical reasoning under real-world constraints.

The aggregation of validated semantic trajectories enables what we term \textit{collective self-evolution}: the emergence of shared diagnostic intelligence exceeding any individual agent's capabilities. This evolution occurs not through centralized retraining on pooled data but through continuous refinement of reasoning strategies based on cross-institutional experience. When an agent successfully navigates a novel presentation to correct a diagnosis, that trajectory becomes available network-wide, enriching the collective repertoire of clinical problem-solving once abstracted and validated.

\section{Architectural Foundations of Decentralized Agentic Systems}
\label{sec:arch_das}

\begin{figure}[t]
    \centering 
    \includegraphics[width=\textwidth]{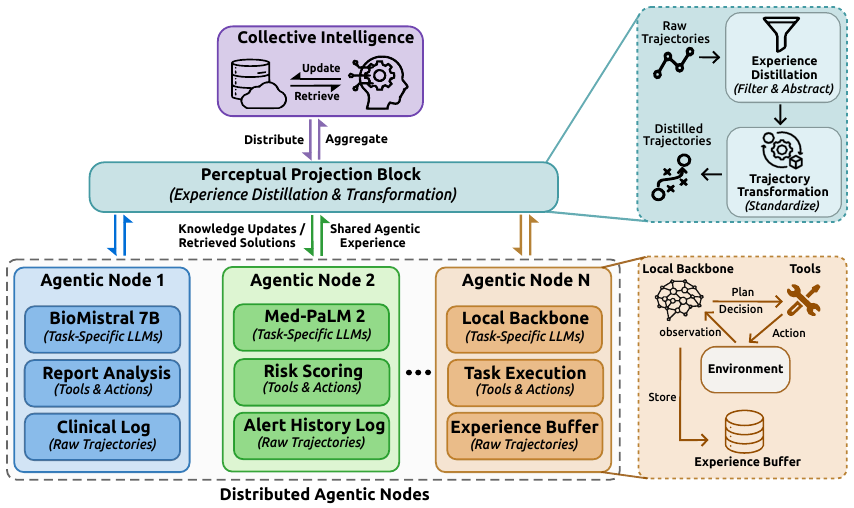}
    \caption{\textbf{Architecture of Decentralized Agentic Systems (DAS).} The framework comprises distributed agentic nodes operating within local clinical contexts, connected through a semantic abstraction layer that transforms raw experiences into standardized reasoning trajectories. The collective intelligence layer aggregates these trajectories to synthesize shared clinical knowledge, which is subsequently distributed back to enhance individual agent capabilities.}
    \label{fig:das-overview}
\end{figure}

We now formalize the DAS architecture. The central challenge is not merely distributing computation but enabling meaningful knowledge synthesis across agents operating in heterogeneous clinical contexts: different patient populations, institutional protocols, specialty focuses, and infrastructure capabilities. As illustrated in Figure~\ref{fig:das-overview}, DAS addresses this through three architectural layers: distributed agentic nodes that generate clinical experiences locally, a semantic abstraction layer that transforms heterogeneous experiences into transferable representations, and a collective intelligence layer that aggregates and redistributes insights across the agent network.

\subsection{Distributed Agentic Nodes}

The foundational unit of DAS is the distributed agentic node, an autonomous agent tailored to local clinical requirements. Formally, we define node $\mathcal{A}_k$ for institution or deployment context $k \in \{1, \ldots, N\}$ as a tuple:
\begin{equation}
    \mathcal{A}_k = \langle \mathcal{M}_k, \mathcal{T}_k, \mathcal{E}_k, \mathcal{C}_k \rangle
\end{equation}
where $\mathcal{M}_k$ denotes the local reasoning backbone (an LLM adapted to the node's clinical specialty and institutional knowledge), $\mathcal{T}_k$ represents the executable tool suite, $\mathcal{E}_k$ is the local experience buffer storing interaction trajectories, and $\mathcal{C}_k$ encodes local context including patient population characteristics, institutional protocols, and regulatory constraints.

Each agentic node operates through an iterative perception-reasoning-action cycle. At each step $t$, the reasoning backbone $\mathcal{M}_k$ receives a clinical observation $o_t$, which may include patient symptoms, diagnostic results, or queries from clinical staff. It generates a reasoning trace $r_t$ explicating its inferential process and selects an action $a_t \in \mathcal{T}_k$ for execution. The environment returns feedback $f_t$, comprising tool outputs, downstream clinical outcomes, or expert corrections. This interaction constitutes an experience tuple:
\begin{equation}
    e_t^k = \langle o_t, r_t, a_t, f_t, c_t \rangle
\end{equation}
where $c_t$ captures relevant contextual metadata, and experiences accumulate as an ordered sequence:
\begin{equation}
    \mathcal{E}_k = \{e_1^k, e_2^k, \ldots, e_T^k\}
\end{equation}

Raw experiences remain locally stored, preserving data sovereignty and enabling compliance with institutional data governance policies.

\subsection{Semantic Abstraction Layer}

As experiences accumulate across distributed nodes, heterogeneity in clinical contexts, observation modalities, and action semantics necessitates transformation before knowledge can be meaningfully shared. We introduce the \textit{Semantic Abstraction Layer}, denoted $\Phi$, implementing a two-stage pipeline: experience distillation and trajectory standardization.

\paragraph{Experience Distillation.} Raw interaction trajectories contain substantial redundancy: routine observations, exploratory actions that yielded no insight, and noise from environmental stochasticity. The distillation function $\phi_{\text{dist}}: \mathcal{E}_k \rightarrow \mathcal{E}_k'$ filters experiences to retain those carrying significant epistemic value:
\begin{equation}
\label{eq:exp-distill}
    \mathcal{E}_k' = \phi_{\text{dist}}(\mathcal{E}_k) = \{e_t^k \in \mathcal{E}_k \mid \mathcal{U}(e_t^k; \mathcal{C}_k) > \delta_k\}
\end{equation}
where $\mathcal{U}(\cdot)$ quantifies epistemic value across dimensions of task success, novelty, error correction, and generalization likelihood. The threshold $\delta_k$ may be context-dependent, reflecting institutional priorities and resource constraints.

The design of $\mathcal{U}(\cdot)$ is an open challenge. We envision hybrid approaches combining automated metrics (e.g., downstream outcome verification) with expert validation, where clinicians flag experiences of particular educational value, analogous to case-based learning in medical education.

\paragraph{Trajectory Standardization.} Following distillation, retained experiences are transformed into a canonical representation enabling cross-context learning. The standardization function $\phi_{\text{std}}: \mathcal{E}_k' \rightarrow \mathcal{S}_k$ maps heterogeneous local experiences into semantic trajectories:
\begin{equation}
\label{eq:traj-standard}
    \mathcal{S}_k = \phi_{\text{std}}(\mathcal{E}_k') = \{\sigma_i\}_{i=1}^{|\mathcal{E}_k'|}, \quad \sigma_i = (o_i^{\text{abs}}, r_i^{\text{abs}}, a_i^{\text{abs}}, f_i^{\text{abs}})
\end{equation}
where each component is abstracted into a domain-general representation. Observations are mapped to standardized clinical concepts (e.g., SNOMED-CT or ICD), reasoning traces are structured according to a common inferential schema, actions are categorized by functional type rather than implementation-specific tool calls, and feedback is encoded as outcome categories with confidence indicators. This standardization enables cross-domain semantic alignment while removing institution-specific identifiers and abstracting patient-level details into population-level patterns.

\subsection{Collective Intelligence through Federated Trajectory Learning}
The collective intelligence layer synthesizes distributed knowledge through \textit{Federated Trajectory Learning}, aggregating standardized semantic trajectories that preserve semantic content rather than averaging model parameters.

Let $\mathcal{S} = \{\mathcal{S}_1, \ldots, \mathcal{S}_N\}$ denote the global trajectory repository, where each $\mathcal{S}_k$ represents standardized trajectories contributed by node $\mathcal{A}_k$:
\begin{equation}
    \mathcal{S}_k = \Phi(\mathcal{E}_k) = \phi_{\text{std}}(\phi_{\text{dist}}(\mathcal{E}_k))
\end{equation}

The collective intelligence mechanism $\mathcal{G}$ updates a shared knowledge representation $\theta_{\text{global}}$:
\begin{equation}
    \theta_{\text{global}}^{(t+1)} = \mathcal{G}\left(\theta_{\text{global}}^{(t)}, \{\mathcal{S}_k^{(t)}\}_{k=1}^{N}\right)
\end{equation}

$\mathcal{G}$ may be instantiated through trajectory-derived insight aggregation, retrieval-augmented reasoning, clinical knowledge graph construction, or distillation into updated reasoning templates, accommodating diverse strategies without prescribing a single formulation. This architecture supports asynchronous learning: nodes contribute trajectories continuously, $\theta_{\text{global}}$ is refined without synchronized participation, and updated knowledge enriches individual $\mathcal{M}_k$ with network-wide insights.

\subsection{Self-Evolution of Distributed Agentic Nodes}
The emergence of collective intelligence creates a reciprocal feedback loop through which individual agentic nodes continuously self-evolve. Formally, node $\mathcal{A}_k$ updates its local reasoning backbone by incorporating knowledge distilled from the global trajectory repository:
\begin{equation}
    \mathcal{M}_k^{(t+1)} = \mathcal{F}\!\left(\mathcal{M}_k^{(t)},\; \mathcal{R}\!\left(\theta_{\text{global}}^{(t)}, \mathcal{C}_k\right)\right),
\end{equation}
where $\mathcal{R}(\cdot)$ is a context-conditioned retrieval function that selects trajectories from $\theta_{\text{global}}$ relevant to the node's local clinical context $\mathcal{C}_k$, and $\mathcal{F}(\cdot)$ integrates retrieved knowledge into existing reasoning capabilities through in-context augmentation \cite{contextaug}, fine-tuning via reinforcement learning from human or AI feedback (e.g., RLHF~\cite{rlhf1, rlhf2}, DPO~\cite{dpo}), or reasoning template refinement.

This self-evolution operates along two complementary axes. \textit{Breadth expansion} occurs as nodes assimilate reasoning trajectories from clinical domains beyond their local experience: a primary care agent encountering a rare cardiac presentation can retrieve and internalize diagnostic pathways developed by specialized cardiology agents elsewhere in the network. \textit{Depth refinement} occurs as nodes receive corrective trajectories that challenge or supersede their existing reasoning patterns, enabling continuous calibration of diagnostic confidence and decision thresholds. As a result, each node converges toward a locally specialized yet globally informed reasoning profile, one that reflects both its accumulated clinical experience and the collective mastery of the distributed network.

\subsection{Deployment Landscape across Healthcare Contexts}
\label{sec:deploy}

\begin{figure}[t]
    \centering 
    \includegraphics[width=\textwidth]{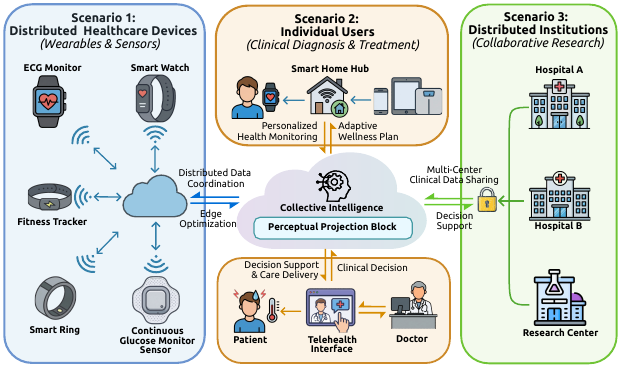}
    \caption{\textbf{Deployment landscape of the DAS framework across healthcare contexts.} The architecture supports deployment across three scales: point-of-care agents providing real-time clinical decision support, institutional agents coordinating departmental expertise, and network-level aggregation enabling multi-center collaborative intelligence.}
    \label{fig:das-deploy}
\end{figure}

The DAS framework supports deployment at multiple scales of clinical organization, as illustrated in Figure~\ref{fig:das-deploy}.

At the \textit{point-of-care} level, agentic nodes operate within clinical workflows (i.e., emergency department triage, outpatient consultation, and surgical planning) providing real-time decision support and accumulating fine-grained experiences from direct clinical interactions.

At the \textit{institutional} level, DAS enables coordination across departments and specialties within a single healthcare organization. A hospital-wide deployment might include specialized agents for cardiology, oncology, primary care, and radiology, each developing domain-specific expertise while contributing to shared institutional knowledge, supporting quality improvement and dissemination of best practices.

At the \textit{network} level, DAS enables secure multi-institutional collaboration among federations of hospitals, academic medical centers, and research institutions while maintaining data sovereignty. This scale is particularly valuable for rare disease diagnosis, where individual institutions encounter insufficient case volumes, and for multi-center clinical research, where distributed agents can identify eligible patients and support protocol adherence.

Across all scales, the semantic abstraction layer ensures knowledge sharing occurs at the level of clinical reasoning rather than raw patient data. We illustrate this through two clinical scenarios: heart failure management across distributed care settings (Box~\ref{box:heart-failure-scenario}) and rare disease diagnosis across a hospital network (Box~\ref{box:rare-disease-scenario}).

\begin{clinicalbox}[box:heart-failure-scenario]{Illustrative Scenario: Distributed Heart Failure Management} 
Consider a network of agents supporting heart failure care across three contexts: a rural primary care clinic, an urban cardiology practice, and a home monitoring system. 

\textbf{Local Experience Generation.} The primary care agent learns to recognize atypical heart failure presentations in patients with limited access to echocardiography, developing expertise in risk stratification using symptoms and basic labs. The cardiology agent accumulates experience with advanced imaging interpretation and medication titration. The home monitoring agent learns to identify early decompensation patterns from daily weight trends and activity data. 

\textbf{Semantic Abstraction.} Each agent's experiences are distilled into transferable trajectories. The primary care agent's identification of heart failure in a patient presenting with fatigue and subtle edema becomes: ``In patients with diabetes and renal impairment presenting with nonspecific fatigue, consider heart failure even without classic dyspnea.'' 

\textbf{Collective Intelligence and Node Self-Evolution.} When the home monitoring agent encounters an ambiguous pattern of modest weight gain with preserved activity levels, it retrieves relevant trajectories from the collective knowledge base and integrates them into its local reasoning backbone, expanding its decompensation recognition beyond patterns observed in its own deployment context. Reciprocally, the primary care and cardiology agents assimilate the monitoring agent's longitudinal decompensation trajectories, refining their own early-warning reasoning. This scenario illustrates how DAS enables knowledge sharing impossible under data-centralization constraints: the rural clinic cannot share patient records with the urban practice, but all three nodes continuously contribute to and self-evolve through collective clinical reasoning. 
\end{clinicalbox}

\section{Governance, Regulation, and Deployment Realities}
\label{sec:govern}

The technical architecture of DAS is necessary but insufficient for real-world adoption. Healthcare AI systems operate within complex governance ecosystems involving institutional policies, regulatory frameworks, professional liability structures, and patient trust relationships. We examine the non-technical dimensions that will determine whether decentralized agentic systems can transition from architectural concept to deployed clinical infrastructure.

\subsection{Regulatory Considerations}

Agentic AI systems in healthcare face an evolving regulatory landscape. In the United States, the FDA's framework for Software as a Medical Device (SaMD) was designed for static algorithms with well-defined inputs and outputs, a paradigm poorly suited to agentic systems that reason dynamically and adapt over time \cite{muhlberger2024regulatory}. Key open questions include whether each institutional agent requires independent regulatory clearance or whether DAS can be evaluated as a system; how continuous learning from trajectory updates interacts with the FDA's predetermined change control plan requirements; and what validation standards apply to collectively derived reasoning patterns emerging from network-level aggregation rather than explicit design.

In the EU, the AI Act classifies most clinical decision-support systems as high-risk, requiring transparency, human oversight, and robustness testing. DAS creates novel compliance challenges: when reasoning reflects contributions from agents across multiple member states, which jurisdiction governs? These questions must be confronted proactively with regulatory bodies.

\begin{clinicalbox}[box:rare-disease-scenario]{Illustrative Scenario: Rare Disease Diagnosis across a Hospital Network} 
Consider a network of 15 academic medical centers, each encountering a small number of patients with hereditary transthyretin amyloidosis (hATTR), a rare, progressive condition frequently misdiagnosed as hypertensive heart disease or idiopathic cardiomyopathy. 

\textbf{Local Experience Generation.} At institution A, a cardiology agent encounters a 65-year-old patient with unexplained left ventricular hypertrophy and bilateral carpal tunnel syndrome. After initially pursuing a standard hypertensive cardiomyopathy workup, the agent, prompted by subtle electrodiagnostic findings, requests a technetium pyrophosphate scan, leading to a correct hATTR diagnosis. At institution B, a neurology agent identifies a pattern linking peripheral neuropathy progression with cardiac biomarker elevation in three patients, eventually connecting the presentation to hATTR. 

\textbf{Semantic Abstraction.} The cardiology agent's reasoning path is distilled into a transferable trajectory: ``In patients with unexplained LVH and concurrent carpal tunnel syndrome or peripheral neuropathy, particularly over age 60, consider technetium pyrophosphate scan to evaluate for cardiac amyloidosis before concluding hypertensive or idiopathic cardiomyopathy.'' The neurology agent contributes a complementary trajectory linking peripheral and cardiac manifestations. 

\textbf{Collective Intelligence and Node Self-Evolution.} When a primary care agent at institution C encounters a patient with progressive neuropathy and mildly elevated troponin, it retrieves these cross-institutional trajectories, flags the possibility of hATTR, and integrates this diagnostic pathway into its local reasoning backbone. Simultaneously, the cardiology and neurology agents at institutions A and B assimilate each other's complementary trajectories, refining their own cross-domain recognition of hATTR. This scenario illustrates how DAS addresses a critical gap in rare disease management: no single institution accumulates sufficient case volume to develop robust diagnostic expertise, but through collective intelligence and reciprocal self-evolution, the network continuously elevates diagnostic capacity across all participating nodes.
\end{clinicalbox}

\subsection{Institutional Governance and Trust Models}

Beyond formal regulation, DAS requires governance structures addressing participation, liability, and reciprocity across three dimensions.

\paragraph{Trajectory curation and quality control.} Who decides which reasoning trajectories enter the collective knowledge base? Uncurated aggregation risks knowledge degradation (see Section~\ref{sec:hallucination}), while overly restrictive curation may stifle the diversity that makes collective intelligence valuable. We envision hybrid models combining automated quality metrics (outcome validation, internal consistency, novelty assessment) with periodic expert review boards analogous to clinical peer review committees.

\paragraph{Liability allocation.} When a recommendation informed by cross-institutional trajectories leads to an adverse outcome, liability is nontrivial. Current frameworks struggle with distributed AI even in non-agentic contexts \cite{price2019medical}. DAS liability models might draw on multi-site clinical guideline precedent, where contributing institutions share collective responsibility for quality while individual providers retain responsibility for application to specific patients.

\paragraph{Participation incentives and fairness.} Institutions must perceive sufficient benefit to justify participation costs, including computational resources, clinical time for quality validation, and governance compliance overhead. Free-rider problems, where institutions consume collective knowledge without contributing proportionally, can erode trust. Contribution-weighted access to collective intelligence and transparent accounting of trajectory utilization may help align incentives. Smaller or less-resourced institutions should not be systematically disadvantaged; enabling rural and community hospitals to benefit from academic medical center expertise is a primary motivation for DAS.

\subsection{Integration with Clinical Infrastructure}

Practical deployment requires integration with EHR platforms. Interoperability standards such as HL7 FHIR provide structured interfaces for clinical observations and decision support. However, institutional variation in EHR configuration, documentation practices, and workflow integration means the \emph{operational context} surrounding each agent differs substantially, affecting experience quality and applicability of received knowledge.

We advocate beginning with standardized tasks: clinical risk scoring, medication interaction checking, or diagnostic workup recommendations, before extending to complex reasoning, enabling iterative validation of trajectory sharing while building institutional trust through demonstrated value.

\section{Technical Challenges and Open Research Directions}
\label{sec:challenges}

The architecture of Section~\ref{sec:arch_das} provides foundational mechanisms for decentralized agentic collaboration, yet several substantial technical challenges remain unresolved, which we examine as follows.

\subsection{Experience Heterogeneity across Clinical Contexts}

Decentralized healthcare systems must reconcile experiences generated across radically different clinical environments. A diagnostic agent in a tertiary academic medical center encounters different patient populations, diagnostic modalities, and institutional protocols than one in a rural primary care clinic. This heterogeneity manifests across three dimensions.

\textit{Statistical heterogeneity} arises from differences in patient demographics and disease prevalence. An agent serving a pediatric oncology unit accumulates experiences with childhood malignancies of limited relevance to one in a geriatric care facility, and direct aggregation across such disparate populations risks distributional biases that degrade diagnostic accuracy.

\textit{Semantic heterogeneity} emerges from differences in clinical vocabulary, documentation practices, and reasoning conventions. The same finding may be described differently by cardiologists and internists; the same conclusion may be reached through different inferential pathways depending on institutional culture. Standardization efforts (as in Equation~\ref{eq:traj-standard}) must navigate the tension between abstraction sufficient for cross-context transfer and preservation of clinically meaningful nuance.

\textit{Operational heterogeneity} reflects differences in available tools, computational resources, and workflow integration. An agent with access to advanced imaging analysis generates fundamentally different trajectories than one limited to laboratory values and clinical notes.

The Semantic Abstraction Layer addresses heterogeneity through trajectory standardization and utility-weighted aggregation. However, determining abstraction levels that enable transfer while preserving clinical validity remains empirically open, and the contribution weights $w_k$ must reflect source-to-target context relevance, requiring either explicit clinical metadata or learned representations of transferability.

\subsection{Hallucination Propagation and Byzantine Resilience}
\label{sec:hallucination}

Large language models exhibit well-documented tendencies toward hallucination \cite{hallu-1, hallu-2, hallu-3}. In agentic systems, this vulnerability extends to entire reasoning trajectories: an agent may construct internally coherent diagnostic pathways based on fabricated associations or nonexistent evidence, and when such trajectories enter the sharing mechanism they risk corrupting collective knowledge network-wide.

This represents a novel form of the classical Byzantine failure problem \cite{byzantine, mlbyzantine}. Agentic hallucination differs critically: hallucinated trajectories may be superficially plausible, evading detection mechanisms designed for obviously erroneous outputs; hallucination propensity may correlate across agents sharing similar architectures, violating independence assumptions underlying Byzantine protocols; and clinical consequences, including incorrect diagnoses and inappropriate treatments, may be severe and difficult to reverse.

The utility threshold in experience distillation (Equation~\ref{eq:exp-distill}) offers partial mitigation but presupposes reliable utility assessment when ground truth outcomes may be delayed. More robust approaches might incorporate multi-agent verification or external validation against established clinical knowledge bases. The governance structures in Section~\ref{sec:govern}, particularly expert trajectory curation, provide a complementary non-technical defense. The optimal balance between automated and human-mediated quality control remains context-dependent, with high-stakes domains such as oncology warranting more stringent validation.

\subsection{Semantic Privacy Leakage from Abstracted Trajectories}
\label{sec:privacy}

A foundational premise of decentralized learning is that sharing derived representations preserves privacy. However, gradient inversion attacks have demonstrated that model updates can leak substantial information about training data \cite{modelinver, gradientinver}, enabling reconstruction of sensitive clinical images from federated exchanges \cite{fedleak, fedapril}.

This vulnerability extends to trajectory sharing. A reasoning path comprising specific tool invocations, diagnostic queries, and decision sequences constitutes a behavioral fingerprint: even without explicit patient identifiers, trajectories querying rare disease databases or following condition-specific diagnostic pathways may reveal diagnoses, genetic predispositions, or lifestyle factors patients expect to remain confidential.

The standardization function provides partial protection, but the tension between utility and privacy persists: sufficiently abstracted trajectories may lose the clinical specificity that makes them valuable. Developing formal privacy guarantees, potentially drawing on differential privacy frameworks adapted to structured reasoning representations, is an important open challenge. The appropriate tradeoff likely varies across contexts, suggesting the need for configurable mechanisms rather than uniform solutions.

\subsection{Coordination Complexity in Asynchronous Multi-Agent Systems}

DAS supports asynchronous operation, allowing agents to contribute trajectories and receive updates without synchronized participation. This flexibility is essential across institutions with different schedules and connectivity constraints, but introduces challenges for coherent collective intelligence.

\textit{Temporal inconsistency} arises when agents operate on different knowledge versions. Recently updated agents generate experiences reflecting current collective understanding, while those with stale knowledge produce trajectories based on outdated patterns, requiring careful versioning and potentially discounting stale inputs.

\textit{Feedback delays} complicate credit assignment. Clinical outcomes validating or refuting diagnostic reasoning may emerge weeks or months after the initial interaction, by which point collective knowledge may have evolved substantially, obscuring attribution of learning signal to specific contributions.

\textit{Coordination overhead} scales with network size. As participating agents grow, mechanisms for trajectory aggregation and knowledge distribution must remain computationally tractable, and ensuring asynchronous updates do not introduce inconsistencies affecting patient care requires careful protocol design and potentially formal verification.

\subsection{Evaluating Collective Intelligence}
\label{sec:eval}

A challenge crosscutting all others is the absence of established evaluation methodologies. Conventional metrics such as accuracy, sensitivity, and specificity apply to individual agents but do not capture the value added by trajectory sharing. We identify three dimensions future work must operationalize.

\paragraph{Marginal improvement from collective knowledge.} Does an agent with access to collective trajectories outperform the same agent in isolation? This requires controlled comparisons measuring diagnostic accuracy and reasoning quality: does the agent consider a broader differential and reach correct diagnoses more efficiently?

\paragraph{Cross-context transfer effectiveness.} Do trajectories from one context (e.g., academic cardiology) improve reasoning in another (e.g., rural primary care)? Transfer effectiveness likely varies across domains, and characterizing this landscape is essential for deployment decisions.

\paragraph{Longitudinal learning dynamics.} Does collective intelligence improve as the trajectory repository grows, or do saturation and diminishing returns emerge? Longitudinal evaluation requires sustained deployment in realistic clinical environments, a significant but essential validation step.

We recommend that early DAS implementations incorporate prospective evaluation protocols from the outset, including pre-registered hypotheses and clearly defined metrics for success and failure.

\section*{Closing Remarks}

The future of healthcare AI lies not in monolithic systems that encode medical knowledge centrally, but in networks of specialized agents learning continuously from clinical experience and sharing insights in privacy-preserving ways. We have proposed Decentralized Agentic Systems as an architectural foundation for this future, one that respects the distributed nature of clinical expertise, the sovereignty of institutional knowledge, and the safety imperatives of patient care. The challenges are substantial, and we have deliberately identified open questions alongside our proposals. The converging pressures of data scarcity, institutional fragmentation, and the maturation of agentic AI capabilities make this a timely moment to chart a path toward collaborative clinical intelligence. The architecture of that collaboration matters, and we believe the principles articulated here: experience over data, reasoning over parameters, and governance alongside technology, point toward a productive direction.

\section*{Funding Statement}
This work was supported by the NGF AiNed Fellowship Grant of A.S.

\bibliography{sn-bibliography}
\end{document}